\documentclass{article}

\PassOptionsToPackage{numbers, compress}{natbib}

\usepackage[preprint]{neurips_2019}

\usepackage[utf8]{inputenc} 
\usepackage[T1]{fontenc}    
\usepackage{hyperref}       
\usepackage{url}            
\usepackage{booktabs}       
\usepackage{amsfonts}       
\usepackage{nicefrac}       
\usepackage{microtype}      
\usepackage{graphicx}
\usepackage{natbib}
\usepackage{amsmath,amsthm,amssymb}
\usepackage{xcolor}
\usepackage{float}

\newcommand{\X}{\mathcal{X}}

\newcommand{\R}{\mathbb{R}}

\DeclareMathOperator*{\argmin}{arg min}

\title{Towards Interpretable Soft Prompts}

\author{
Oam Patel$^{*}$ \\
Harvard University
\And
Jason Wang$^{*}$ \\
Harvard University
\And
Nikhil Shivakumar Nayak$^{*}$ \\
Harvard University
\AND
Suraj Srinivas \\
Harvard University
\And
Himabindu Lakkaraju \\
Harvard University
}

\begin{document}

\maketitle
\def\thefootnote{*}\footnotetext{These authors contributed equally to this work. Correspondence to: Oam Patel <opatel@college.harvard.edu>, Jason Wang <jasonwang1@college.harvard.edu>, Nikhil Shivakumar Nayak <nnayak@g.harvard.edu>.}\def\thefootnote{\arabic{footnote}}






\vspace{-1em}

\begin{abstract}

Soft prompts have been popularized as a cheap and easy way to improve task-specific LLM performance beyond few-shot prompts. Despite their origin as an automated prompting method, however, soft prompts and other trainable prompts remain a black-box method with no immediately interpretable connections to prompting. We create a novel theoretical framework for evaluating the interpretability of trainable prompts based on two desiderata: faithfulness and scrutability. We find that existing methods do not naturally satisfy our proposed interpretability criterion. Instead, our framework inspires a new direction of trainable prompting methods that explicitly optimizes for interpretability. To this end, we formulate and test new interpretability-oriented objective functions for two state-of-the-art prompt tuners: Hard Prompts Made Easy (PEZ) and RLPrompt. Our experiments with GPT-2 demonstrate a fundamental trade-off between interpretability and the task-performance of the trainable prompt, explicating the hardness of the soft prompt interpretability problem and revealing odd behavior that arises when one optimizes for an interpretability proxy.

\end{abstract}

\section{Introduction}

\begin{figure}[h]
    \centering
    \includegraphics[width=0.79\textwidth]{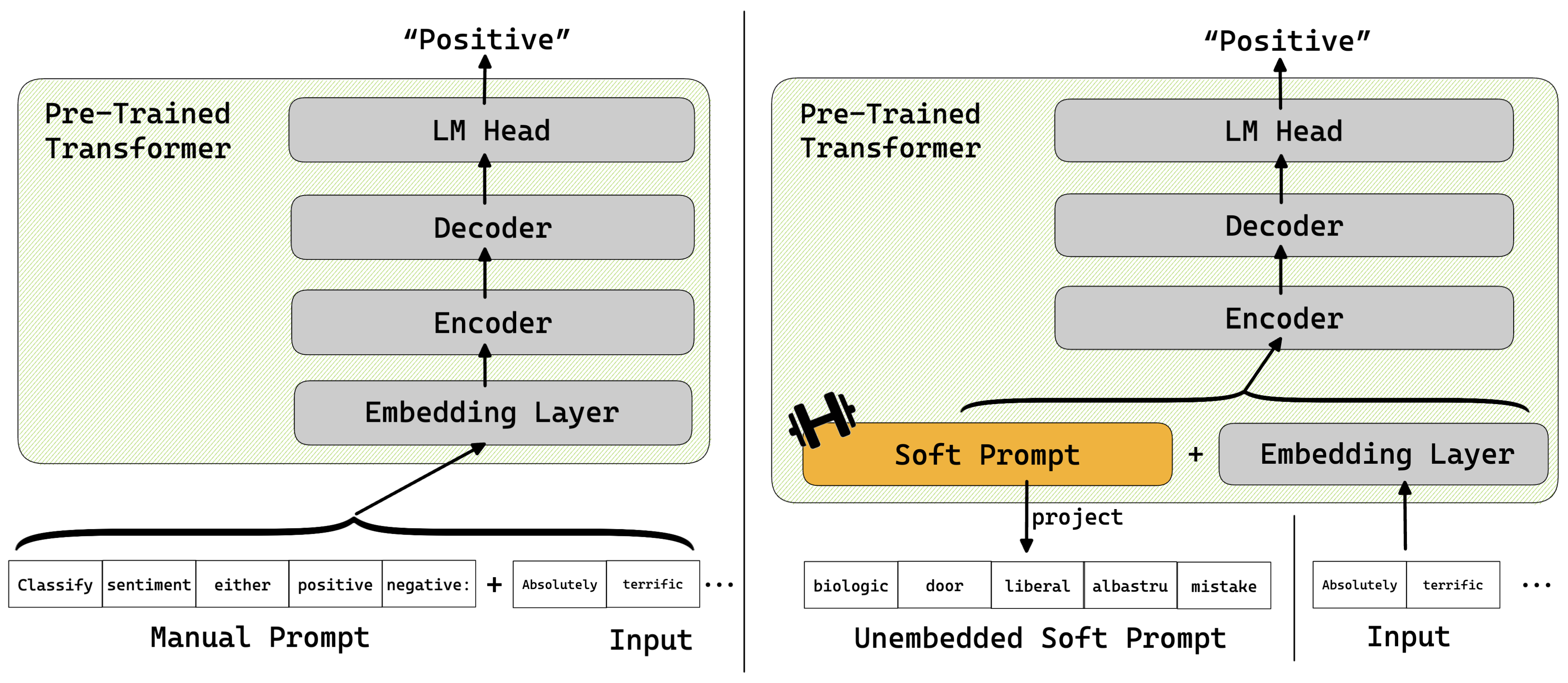}
    \caption{Soft Prompting Illustrated}
    \label{fig:soft_prompt}
\end{figure}

As large language models (LLMs) continue to scale and their emergent zero and few-shot abilities become more capable, prompting (the way that you construct your input to the LLM for a desired output) has grown in importance relative to traditional transfer learning or fine-tuning as an alternative learning paradigm that increases the flexibility of a model to be used for different tasks. A recent area of study \cite{lester2021power} has looked into \textit{soft prompting}, a technique where instead of fine-tuning the weights of the transformer for the task, the transformer's weights are frozen and the model learns the best input prompt for the task, interpolating continuously within the latent space of the text embeddings (see Figure \ref{fig:soft_prompt}).

Unfortunately, soft prompts are uninterpretable \cite{lester2021power}. Given their position as activations prepended to the input task and the convenient fact that they lie in the same space as the token embedding dimension, it is tempting to consider them analogous to prompts. After all, a small subset of the space could be simulated by manual prompting. However, unembedding the soft prompts (choosing the closest discrete tokens to the continuous embedding) results in inscrutable gibberish. Furthermore, prepending the unembedded soft prompt into the transformer loses all the performance gain of the continuous soft prompt embedding. Therefore, soft prompts currently serve as a black-box method more analogous to constrained fine-tuning than to clever prompting.

There are a number of possible reasons for why soft prompts are uninterpretable. First, the embedding space is very large, so the token vectors are scattered sparsely in the space. This means that there is with high probability a large distance between any soft prompt embedding and the nearest vocabulary vector. Thus, there is a relatively large loss of information when unembedding, and so it is expected that the unembedded prompt will not have the same performance boost as that of the soft prompt. Another potential issue is that the transformer only ever sees the discrete vocabulary tokens during training, so it never learns about the continuous space in between token embeddings. Therefore, the soft prompt space may be expected to be highly jagged and unruly, so any interpolation between token embeddings may be highly nonlinear and would probably be inconceivable.

Nonetheless, there is strong incentive to make soft prompts interpretable. Given their superior performance to few-shot prompts and other manually crafted prompts, soft prompts might be able to tell us how to create better prompts. It can only do that if we can parse the soft prompt. Moreover, interpretable soft prompts could provide insight into the model's reasoning or comprehension of the task. Our research question is thus,

\vspace{-0.3em}
\begin{center}
\textit{    How can we make the content of automated prompts more interpretable?
}\end{center}
\vspace{-0.3em}

Our paper creates a theoretical framework for thinking about this question. Most importantly, we explore the feasibility of jointly optimizing for task performance and interpretability and whether there exists some inherent trade-off between interpretability and task performance.

\section{Related Work}

Transformers have proven to be a very versatile architecture for language modeling \cite{vaswani2017attention}, and soft prompts have recently emerged as a way to tune prompts for languange models by optimizing over embedding space instead of token space (which is discrete) \cite{lester2021power}. This method has proven to be a sample-efficient alternative to finetuning that more naturally ``elicits" the knowledge inside models.  Unlike traditional fine-tuning where a specific target sequence is provided as the prompt, soft prompts provide a partially masked input sequence where the model is encouraged but not required to predict specific tokens. Plus, soft prompts don't actually alter the model's weights and thus maintain the ability to generate diverse outputs. Unfortunately, however, these soft prompts after prompt tuning are often uninterpretable as the embedding space is high-dimensional and there's no smoothness constraints. There has been little work on analyzing the interpretabilty of soft prompts in-depth with most claims being qualitative and example-driven \cite{lester2021power}. 

Soft prompts are a natural extension to previous `gradient-guided' but ultimately discrete optimization methods such as AutoPrompt \cite{shin2020autoprompt}. AutoPrompt is designed for masked-language models and works by greedily filling in prompt tokens to provide the largest increase in likelihood of the target label. There has also been work on more efficiently solving the discrete optimization problem, including up to applying RL as in RLPrompt \cite{deng2022rlprompt}, which we describe in more detail in the methods section. So far, these methods have not engaged in-depth with the interpretability concerns of prompt tuning. The discrete prompts found by both AutoPrompt and RLPrompt tend to have completely unrelated tokens or even unexpected tokens that don't occur frequently in the training corpus indicating that the prompting procedure is doing something unexpected. 

In terms of explicitly addressing interpretability, iPrompt \cite{singh2022explaining} is a follow-up work to AutoPrompt that tries to improve the coherence of the prompt by starting with proposal prompts and then reranking and mutating the prompts to convergence in an evolutionary-like process. Bulat et al. \cite{bulat2022language} actually does look at soft prompts, adding a regularization term to make unembedded soft prompts more closely resemble manually constructed prompts. While reasonably effective, both of these methods are constraining insofar as it requires us to already have manually constructed prompts. In fact, we'd most like to be able to use soft prompts where we are unable to construct prompts that effectively elicit the behavior we'd like. Our proposed method explores this harder regime of optimization without the crutch of manually constructed prior prompts.


\section{Preliminaries}



The formal problem setup is as follows. Denote $\X$ to be the space of possible tokens (the known vocabulary) and consider a typical text-to-text autoregressive encoder-decoder transformer $f_\theta:\X^n\to\X^m$ with the following notation for the components:
\begin{itemize}
    \item A linear embedding $f_{V}:\X\to\R^d$ that maps input tokens $x_1,\dots,x_n\in\X$ to vector representations $\bar{x}_1,\dots,\bar{x}_n\in\R^d$
    \item An encoder $f_{enc}:\R^{n\times d}\to\R^{n\times d}$ mapping these vector representations $\bar{x}_1,\dots,\bar{x}_n\in\R^d$ to latent representations $\bar{z}_1,\dots,\bar{z}_n\in\R^d$
    \item A decoder $f_{dec}:\R^{(t-1)\times d}\times\R^{n\times d}\to\R^{d}$ mapping previous decodings $\bar{y}_1,\dots,\bar{y}_{t-1}$ and the latent representations $\bar{z}_1,\dots,\bar{z}_n\in\R^d$ to an output vector representation $\bar{y}_t\in\R^d$
    \item A linear language modeling head $f_{head}:\R^d\to\R^{|\X|}$ that maps decoder output vector $\bar{y}_t$ to vocab logits $l_1,\dots,l_{|\X|}$ defining the output conditional distribution from which $y_t$ is sampled. A decoding scheme (e.g. beam search) autoregressively calls the decoder and LM head to give $y_1,\dots,y_n$.
\end{itemize}
Prompting is the act of prepending prompt tokens $P_1,\dots,P_L\in\X$ to the input tokens before passing the concatenated result into the encoder: 
$$\bar{z}_{P_1},\dots,\bar{z}_{P_L},\bar{z}_1,\dots,\bar{z}_n=f_{enc}(f_{V}(P_1),\dots,f_{V}(P_L),f_{V}(x_1),\dots,f_{V}(x_n))$$

Soft prompting is essentially a trainable prompt. Given a pre-trained $f_\theta$ with frozen weights, add a trainable parameter $\bar{P_1},\dots,\bar{P_L}\in\R^d$ with respect to the particular downstream task such that
$$\bar{z}_{P_1},\dots,\bar{z}_{P_L},\bar{z}_1,\dots,\bar{z}_n=f_{enc}(\bar{P_1},\dots,\bar{P_L},f_{V}(x_1),\dots,f_{V}(x_n))$$

Generally, $|\X|\ll\R^d$ so $f_{V}$ is not invertible, but we can define the next best thing $f_{U}:\R^d\to\X$ s.t. $f_{U}(\bar{x})=\argmin_{y\in\X}D(f_{V}(y),\bar{x})$ based on some distance metric $D$ such as Euclidean or cosine distance. This gives us a way to ``unembed" the soft prompt $\bar{P}_1,\dots,\bar{P}_L$ back into the nearest vocabulary tokens $P_1,\dots,P_L$.

\section{Problem Statement \& Methodology}

The key issue and research question is that soft prompts are not interpretable. Specifically, if we have learned the soft prompt $\bar{P}_1,\dots,\bar{P}_L$, then $f_{U}(\bar{P}_1),\dots,f_{U}(\bar{P}_L)$ is inscrutable to the human eye. Getting interpretable soft prompts would give us a better idea (in human understandable terms) of the loss landscape of prompts, illuminating potential directions for adjustment or otherwise inform human-designed prompts. They may provide insight into the model's reasoning or comprehension of the task and may better highlight human-understandable patterns or quirks of prompt behavior that appear in the model that could point to interesting underlying mechanisms. With this motivation in mind, we establish two desiderata for automated prompt generation:
\begin{enumerate}
    \item \textit{Faithfulness.} A prompt is faithful if unembedding the prompt accurately captures the essence of the embedded prompt. If unembedding the prompt results in an interpretable sentence but does not actually reflect the soft prompt, it is of no use to us since any conclusions we make from the unembedded prompt would not transfer to the actual prompt. The spirit of faithfulness is hard to define, but we consider three natural distance metrics for gauging a prompt's faithfulness (where lower distance indicates higher faithfulness):
    \begin{itemize}
        \item Distance Differential. If the distance between the soft prompt and the unembedded soft prompt is small, we can consider the model to be faithful. In other words, let $\Delta_{D}(\bar{P}_1,\dots,\bar{P}_L)=\frac{1}{L}\sum_{i=1}^L\min_{y\in\X}D(f_{V}(y),\bar{P}_i)^2$ for some distance metric $D$ (Euclidean, cosine distance, etc.).
        \item Output Differential. If the generated output of the model when given a soft prompt vs. unembedded prompt is small, we can consider the model to be faithful. This can be done between the logits: let $l_{1_t},\dots,l_{|\X|_t}$ be the logits for the $t$th output token for the soft prompted model and let $l'_{1_t},\dots,l'_{|\X|_t}$ be the logits for the $t$th output token for the unembedded soft prompted model. Then $\Delta_{Output}(\bar{P}_1,\dots,\bar{P}_L)=\frac{1}{L}\sum_{i=1}^L(\frac{1}{|\X|}\sum_{j=1}^{|\X|}(l_{j_i}-l'_{j_i})^2)$. Alternatively, we can compare just on the most probable token for a looser notion of faithfulness.
        \item Performance Differential. If the performance on the downstream task is similar between using the soft prompt or manually prompting with the unembedded soft prompt, we can consider the model to be faithful. In other words, if $P_i=f_{U}(\bar{P}_i)$, then let $\Delta_L(\overline{P}_1,\dots,\overline{P}_L)=|L(f_\theta(x_1,\dots,x_n|\overline{P}_1,\dots,\overline{P}_L))-L(f_\theta(P_1,\dots,P_L,x_1,\dots,x_n))|$ for some loss function $L$ (e.g., accuracy).
    \end{itemize}
    \item \textit{Scrutability}. A prompt is scrutable if unembedding the prompt results in utterance that is human-interpretable. Previous work in automated prompt generation has opted for grammatical rules or templates with cloze rules \cite{shin2020autoprompt}, but this limits expressivity, and distance to a grammar can be complex to compute in a differentiable manner. We opt for defining scrutability with respect to the well-established perplexity metric evaluated on pre-trained GPT2-XL. If $P_i=f_{U}(\bar{P}_i)$, then $\mathrm{Perplexity}(P_1,\dots,P_L)=\exp(\frac{1}{L}\sum_{i=1}^L-\log p_{\phi}(P_i|P_{<i}))$ where $p_\phi$ is the GPT2-XL language model's output conditional distribution. Perplexity is the (exponential of) average negative log-likelihood of the prompt predicted by GPT2-XL, which represents how likely the prompt would appear in natural human language, viewing GPT2-XL as a proxy for elusive ground truth. Lower perplexity indicates greater human scrutability.
\end{enumerate}

These two desiderata leads us to consider discrete optimization (or close to it) and regularization with perplexity. Faithfulness distance is zero when the prompt embeddings are discrete token embeddings ($\bar{P}=P$), as these prompts would be fully simulatable with manual prompts. While this is a severe constraint on the space of the trainable prompt, the trainable prompt still has immense expressive capability combinatorially ($|\mathcal{X}|^{L}$ potential prompts). Given that neither soft prompts nor discrete prompts \cite{lester2021power,deng2022rlprompt} have inherent scrutability, it seems that for the trainable prompt's content to be interpretable, we must provide explicit incentive to learn a scrutable prompt. Using our metric of perplexity as a proxy for scrutability, we can add a regularization or reward term for it in any optimization scheme, with the weight of term (the importance of interpretability) controllable with a hyperparameter. In particular, we experiment with two methods:

\begin{enumerate}

    

    \item Hard Prompts Made Easy (PEZ). A quick approach to constrained optimization with is PEZ \cite{wen2023hard}. The PEZ algorithm is particularly useful when the constraints on the optimization problem are non-linear or difficult to express explicitly. Like proximal gradient descent, PEZ is an iterative algorithm that starts from an initial guess and refines the solution iteratively. We keep a continuous embedding, project to the nearest token embedding vector, calculate the gradient with respect to the projection, but then apply the gradient to the continuous embedding. They claim that this has better accuracy and has less of a tendency to get stuck in local minima than alternative methods such as projected gradient descent. The PEZ algorithm consists of iteratively updating the parameters $\bar{P}_i$ of the soft prompt as follows: $$\bar{P}_i^{(t+1)} = \bar{P}_i^{(t)} - \alpha \nabla_{\mathrm{Proj}(\bar{P}_i^{(t)})} L(\mathrm{Proj}(\bar{P}_i^{(t)}))$$
    
    where $t$ is the iteration number, $\alpha$ is the learning rate, $L$ is the task loss, and $\mathrm{Proj}(\cdot)$ is the projection operator that projects the soft prompt onto the closest vocab embedding to satisfy faithfulness.
    
    \item Discrete Optimization (RLPrompt). We take the best existing discrete prompt optimizer that is unconstrained by templates or grammar (as to be match flexiblility and expressiveness of soft prompts) called RLPrompt \cite{deng2022rlprompt}. At a high level, RLPrompt trains a policy network to generate prompts. The policy network is a frozen LLM with a trainable MLP just before the LM head, which gives us most of the expressivity of a LLM but for just a small number of trainable parameters. It is trained using soft Q-learning with reward stabilization where the reward is the accuracy on the downstream task. We'll add $\alpha$ times the perplexity of the prompt as an additional regularizer to hopefully get more interpretable prompts at the end. 
\end{enumerate}

For PEZ, we backpropagate through a frozen GPT2-XL through unembedding using the same projection algorithm used in PEZ. The function we want to optimize is the joint objective function consisting of the task loss and the interpretability regularization term (perplexity). Let us denote the parameters of GPT2-XL as $\phi$ and the task loss as $L(\cdot)$. The joint objective function can then be written as: $$J(\bar{P}_i) = L(\bar{P}_i)+\lambda\mathrm{Perplexity}_\phi(f_{U}(\bar{P}_1),\dots,f_{U}(\bar{P}_L))$$

where $\lambda$ is a hyperparameter that controls the importance of the interpretability regularization term. The gradient of the joint objective function can be computed by chain rule, using PEZ:
$$\nabla_{\bar{P}_i} J(\bar{P}_i) = \nabla_{\bar{P}_i}L(\bar{P}_i)+\lambda\nabla_{f_{U}(\bar{P}_i)}\mathrm{Perplexity}_\phi(f_{U}(\bar{P}_1),\dots,f_{U}(\bar{P}_L))\nabla_{\bar{P}_i}f_{U}(\bar{P}_i)$$
but $\nabla_{\bar{P}_i}f_{U}(\bar{P}_i)$ is intractable so we use the PEZ approximation but on $f_{U}$ instead of $\mathrm{Proj}$ (``Unembedded Gradient Descent") by calculating
$$\nabla_{f_{U}(\bar{P}_i)} J(f_{U}(\bar{P}_i)) = \nabla_{f_{U}(\bar{P}_i)}L(f_{U}(\bar{P}_i))+\lambda\nabla_{f_{U}(\bar{P}_i)}\mathrm{Perplexity}_\phi(f_{U}(\bar{P}_i),\dots,f_{U}(\bar{P}_i)$$
with update rule $$\bar{P}_i^{(t+1)} = \bar{P}_i^{(t)} - \alpha \nabla_{f_{U}(\bar{P}_i^{(t)})} J(f_{U}(\bar{P}_i^{(t)}))$$


\section{Experimental Results}

We measure the results of baseline softprompts, PEZ, and RLPrompt with perplexity regularization on different datasets.\footnote{The code for our experiments can be found at \url{https://github.com/NikhilNayak-debug/towards_interpretable_softprompts}.} We expect to see an appreciable difference in perplexity between the PEZ and RLPrompt methods compared to baseline softprompts which would be inline with qualitative results from previous work and provide a step towards more interpretable soft prompts. 

\begin{figure}[h]
    \centering
    \includegraphics[width=0.5\textwidth]{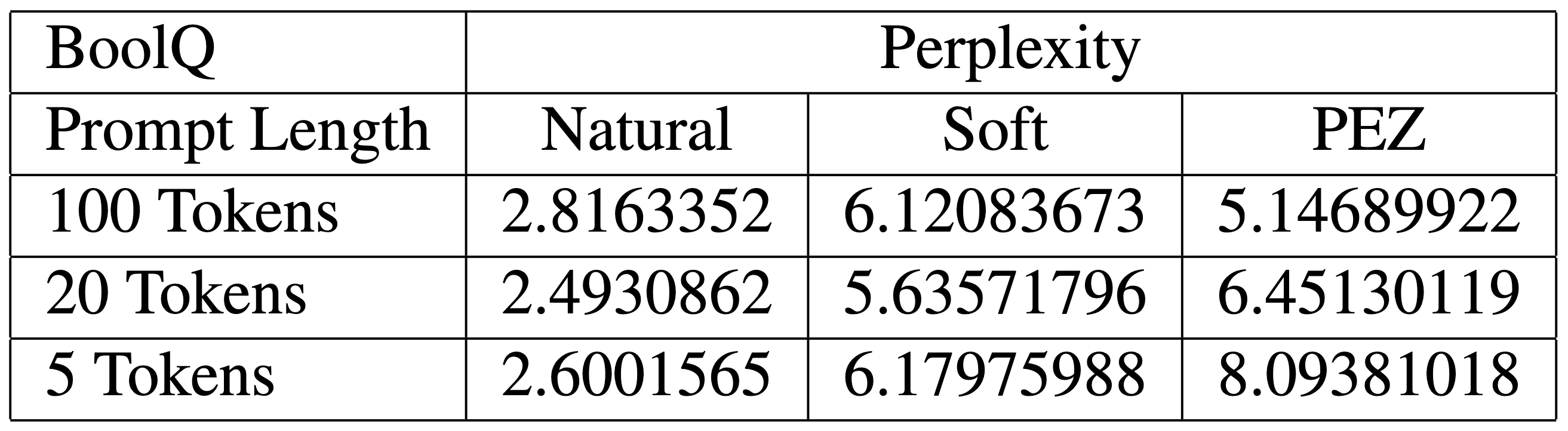}
    \caption{Soft Prompt and PEZ on BoolQ}
    \label{fig:boolq}
\end{figure}

\textbf{Soft Prompt and PEZ:} We use the T5-small architecture in line with the original soft prompt paper \cite{lester2021power} and test on the BoolQ question and answering dataset \cite{clark2019boolq}. This dataset consists of yes/no reading comprehension questions given a passage. To evaluate T5-small on this dataset, we concatenate the passage and question as the input and then compare the logits of the `yes" and `no" tokens of the first output token, taking the larger one as the answer. The best accuracy of the soft prompt over 5 trials was 62\% for the 20 and 100 token prompts, but for PEZ the same was 38\% (same as no prompt), intimating a failure of the PEZ method to converge. Thus, we did not pursue regularizing PEZ with perplexity. Perplexity results for soft prompts and PEZ can be found in Figure \ref{fig:boolq}.

\textbf{RLPrompt:} For RLPrompt, we add a reward term for perplexity. In our setting, we learn prompts for DistilGPT2 (the base model) and use GPT2-XL for perplexity calculation. We use a larger model for perplexity calculation just because we expect GPT2-XL's perplexity measure to more strongly correlate with human ratings of interpretability. We evaluate RLPrompt on two settings: sentiment classification and text style transfer. 

\begin{figure}[h]
    \centering
    \includegraphics[scale=0.37]{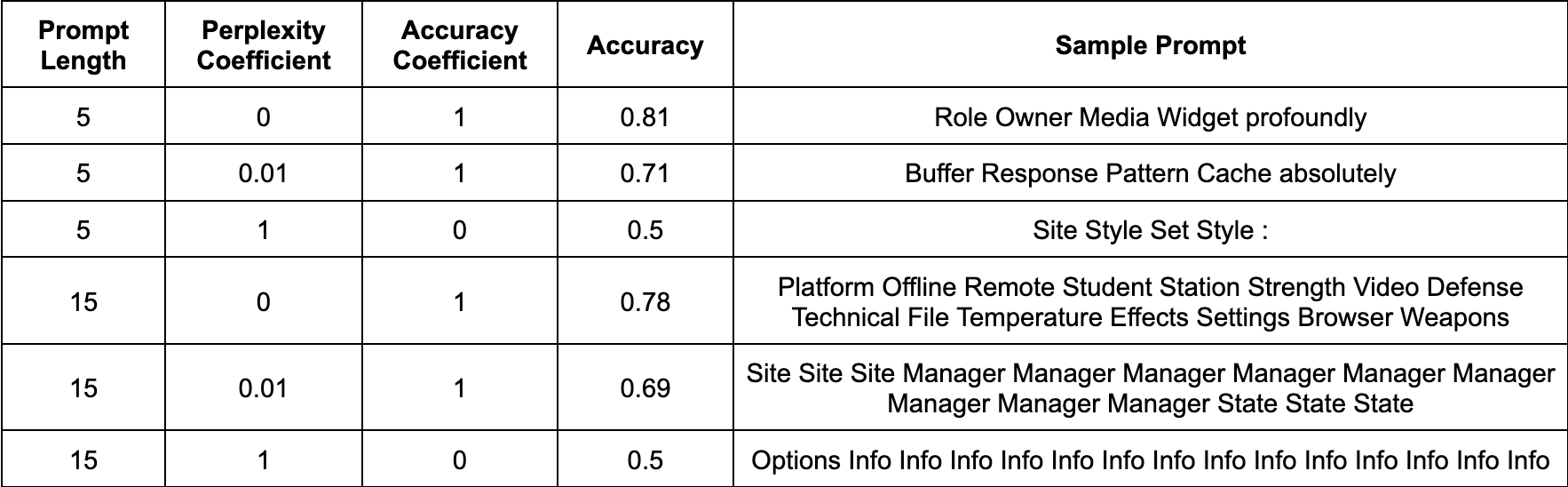}
    \caption{Overview of RLPrompt with Perplexity Regularization on SST-2.}
    \label{fig:rlprompt}
\end{figure}

Few-shot sentiment classification is the task of predicting the sentiment of a piece of text given only a small number of labeled examples. The SST-2 dataset \cite{socher2013recursive} is a popular benchmark for this task, consisting of movie reviews labeled as either positive or negative. We use the binary cross entropy loss along with alpha times the perplexity of the as the reward signal in this setting. Alpha is a hyperparameter that controls the weighting of these two loss terms. It is observed that as alpha increases, the perplexity of the generated prompt decreases, but at the cost of a decrease in accuracy on the task. However, even with higher alpha values that reduce perplexity, the generated prompt remains largely disconnected from the task context and is inconsistent across training epochs, despite the increased prevalence of common English words, which strangely are mostly computer science themed. Figure \ref{fig:rlprompt} illustrates the trade-off between accuracy and perplexity along with sample prompts for each of the experiments. 

\begin{figure}[h]
    \centering
    \includegraphics[scale=0.3]{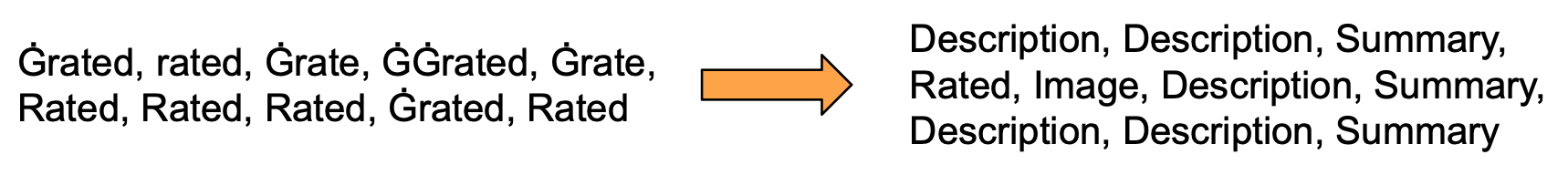}
    \caption{Style transfer prompt with $\alpha = 0$ perplexity regularization on the left, $\alpha = 0.5$ on the right.}
    \label{fig:bestrlprompt}
\end{figure}

Text style transfer involves taking a piece of text and translating it to a different style while preserving the content of the text. For example, we might take a review that is written angrily but has important content about the location and translate it into a helpful style. In this setting, we consider text style transfer on the Yelp reviews dataset \cite{shen2017style}, which comes with classifiers that score the quality of translation and quality of style. We use these terms along with $\alpha$ times the perplexity of the prompt as a reward signal. Our takeaway is that regularizing by perplexity gives us much lower perplexity prompts (as expected) with only a minimal hit in overall content score (about 10\%). Unfortunately, lower perplexity only roughly translated to more interpretable prompts in our setup. Figure \ref{fig:bestrlprompt} shows that we go from unintelligible, low-probability toknes like `Ggrated' to more semantically related, higher probability tokens like `description'. We don't find any consistency in the prompt across phrases though. It's more that individual words are more decipherable. 

We experiment with varying levels of $\alpha$ and report our results in Figure 6. Again, we find a small hit to overall performance (content and style scores) and find that our perplexity goes down as expected). There is a fine line here since if we turn up $\alpha$ to be $1$, the RL algorithm hones in on just optimizing for perplexity and makes no progress on the style or content front. 

\section{Conclusion}

In this work, we develop a framework consisting of two desiderata, faithfulness and scrutability, that interpretable trainable prompts should possess. We used these desiderata to inform our experiments on discrete prompt tuning methods PEZ and RLPrompt with the addition of a perplexity reward term. We tested these two methods on a variety of different tasks (question answering, sentiment analysis, and text style transfer) for GPT-2. From varying the weight of the perplexity reward term, we find that there is a trade-off between scrutability and performance, as well as faithfulness and performance. Analyzing the prompts tells us that such direct optimization of perplexity still does not reliably result in qualitatively scrutable results (our most successful scrutable prompt was with text style transfer, but was not scrutable for the sentiment analysis task). This means that perplexity might be a flawed proxy for scrutability (in which case we could test metrics like BLEU or MAUVE). Alternatively, it may be that soft prompts cannot be meaningfully interpreted in the framing of a prompt.

Future work may also look into the interpretability of adversarial prompts that are optimized to minimize rather than maximize task performance. Another desirable, although not strictly necessary, attribute might be the ability to generalize across different models and scales. Even larger language models \cite{lester2021power} than the ones tested here have shown emergent abilities that scale better with soft prompts. This project initially started with the idea of regularizing the latent space of the embeddings with the variational trick \cite{kingma2013auto,ferner2022benefits}, which still might be an interesting idea to test. Finally, another interesting avenue of future research may turn away from scrutability but instead consider the hypothesis that soft prompts aggregate signals from the entire training dataset. It would be interesting to correlate soft prompts to training examples via influence functions and interpret them in context with few-shot prompts, whether the most influential n-shot does better than the least influential n-shot.

\section{Author Contributions}
Oam Patel and Jason Wang came up with the problem statement. Oam Patel and Jason Wang came up with the key ideas of the main methodology. Oam Patel, Jason Wang, and Nikhil Nayak implemented the main methodology. Oam Patel and Jason Wang led the development of theoretical results along with inputs from Nikhil Nayak. Nikhil Nayak also identified appropriate datasets and Oam Patel and Jason Wang carried out the necessary preprocessing of these datasets. Oam Patel, Jason Wang, and Nikhil Nayak designed and implemented the experimental evaluation. Jason Wang led the writing of the final project report along with inputs from Oam Patel and Nikhil Nayak. Oam Patel, Jason Wang, and Nikhil Nayak documented the code and created scripts for running the entire pipeline with just a handful of commands. All the authors also helped each other with brainstorming as well as double-checking their work.

\bibliographystyle{abbrvnat}
\nocite{*}
\bibliography{final}

\appendix 

\section{Appendix: Additional Figures}

\begin{figure}[H]
    \centering
    \includegraphics[width=0.9\textwidth]{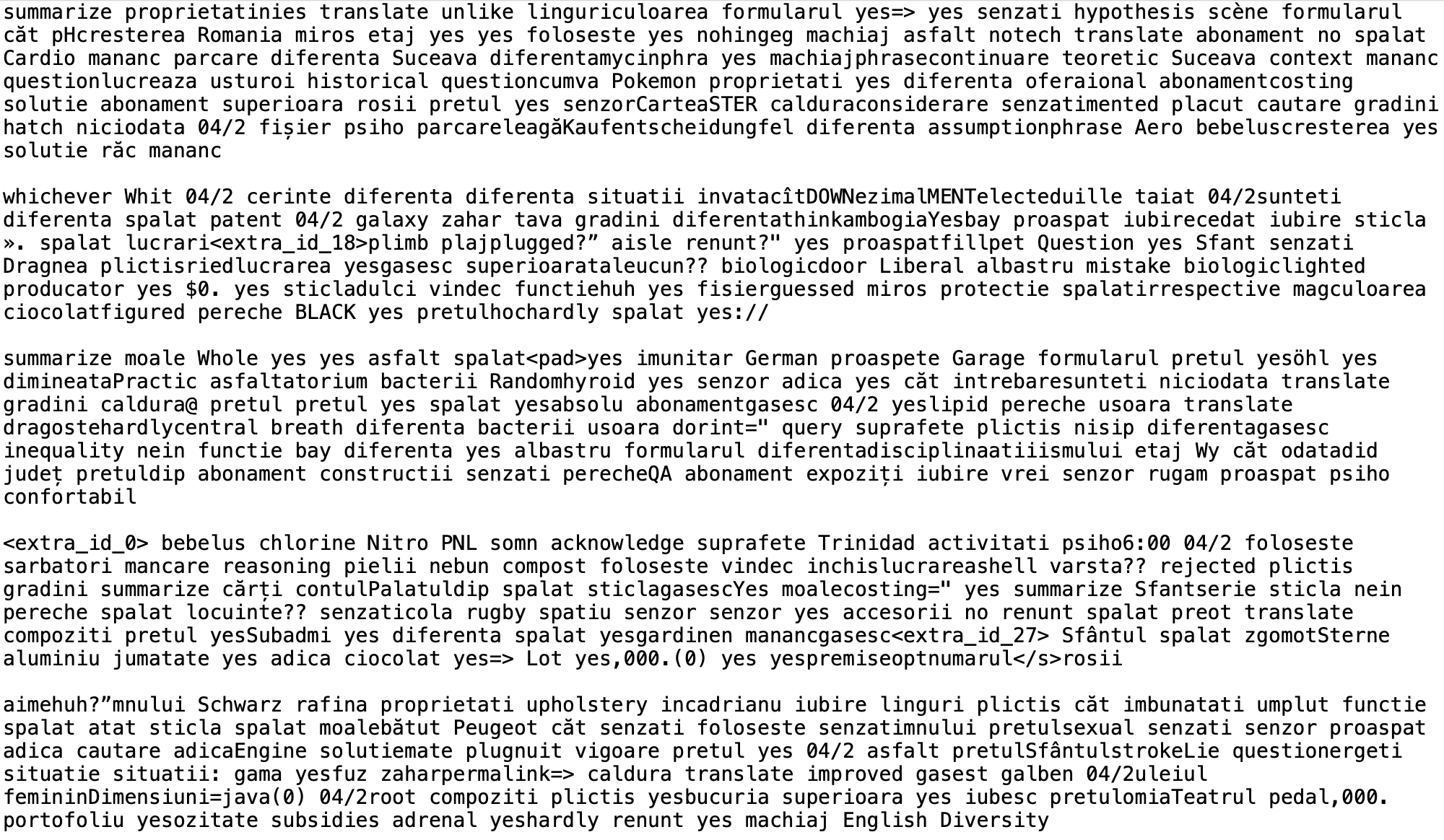}
    \caption{100 Token Soft Prompts for BoolQ}
    \label{fig:baseline}
\end{figure}

\begin{figure}[H]
    \centering
    \includegraphics[scale=0.4]{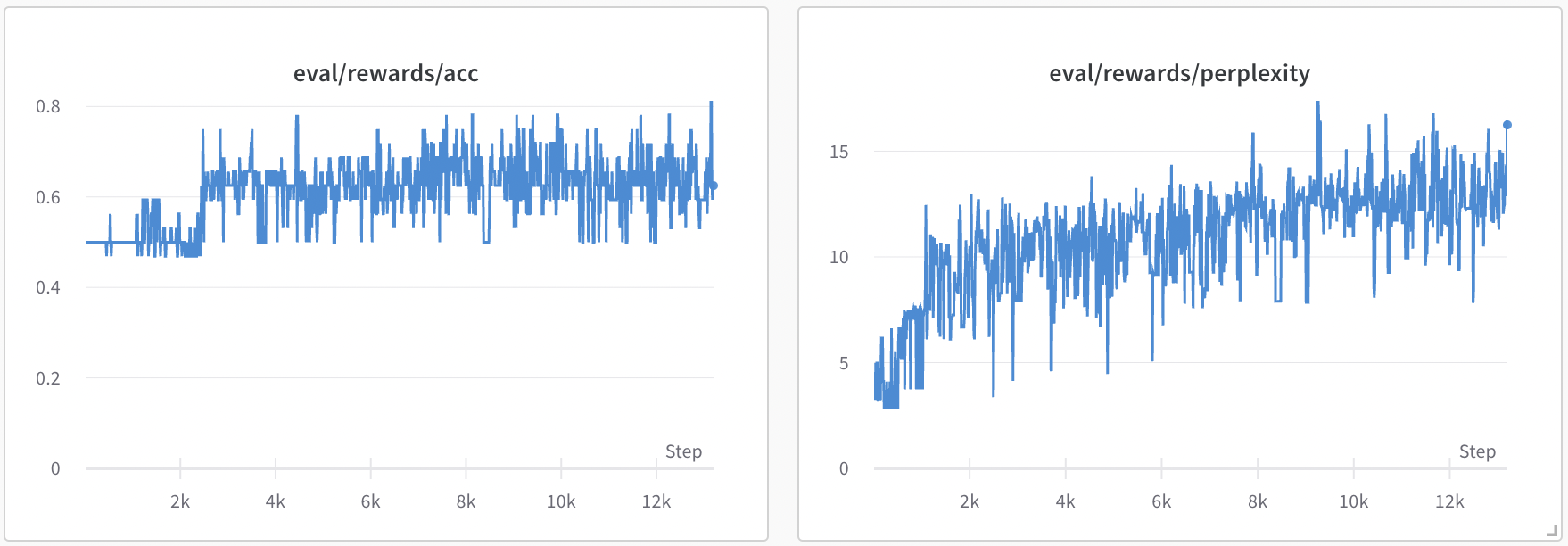}
    \caption{Accuracy and perplexity variation with training steps using only task loss as reward.}
    \label{fig:my_label}
\end{figure}

\begin{figure}[H]
    \centering
    \includegraphics[scale=0.4]{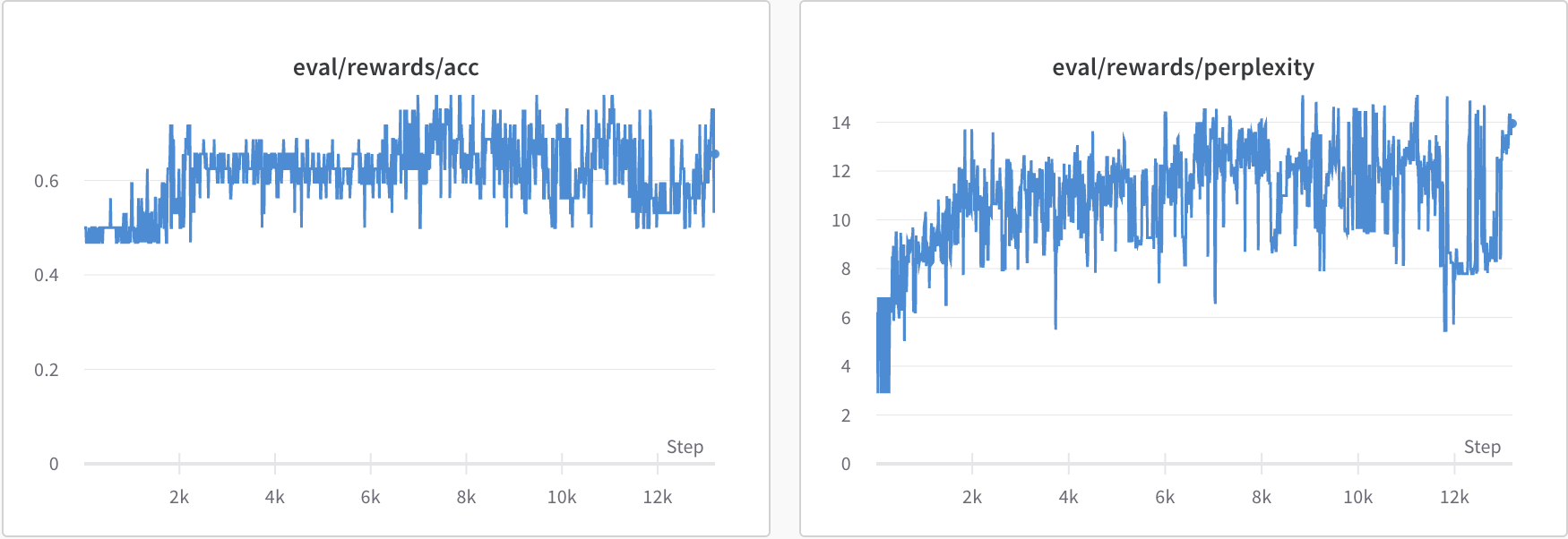}
    \caption{Accuracy and perplexity variation with training steps using task loss and perplexity as reward.}
    \label{fig:my_label}
\end{figure}

\begin{figure}[H]
    \centering
    \includegraphics[scale=0.4]{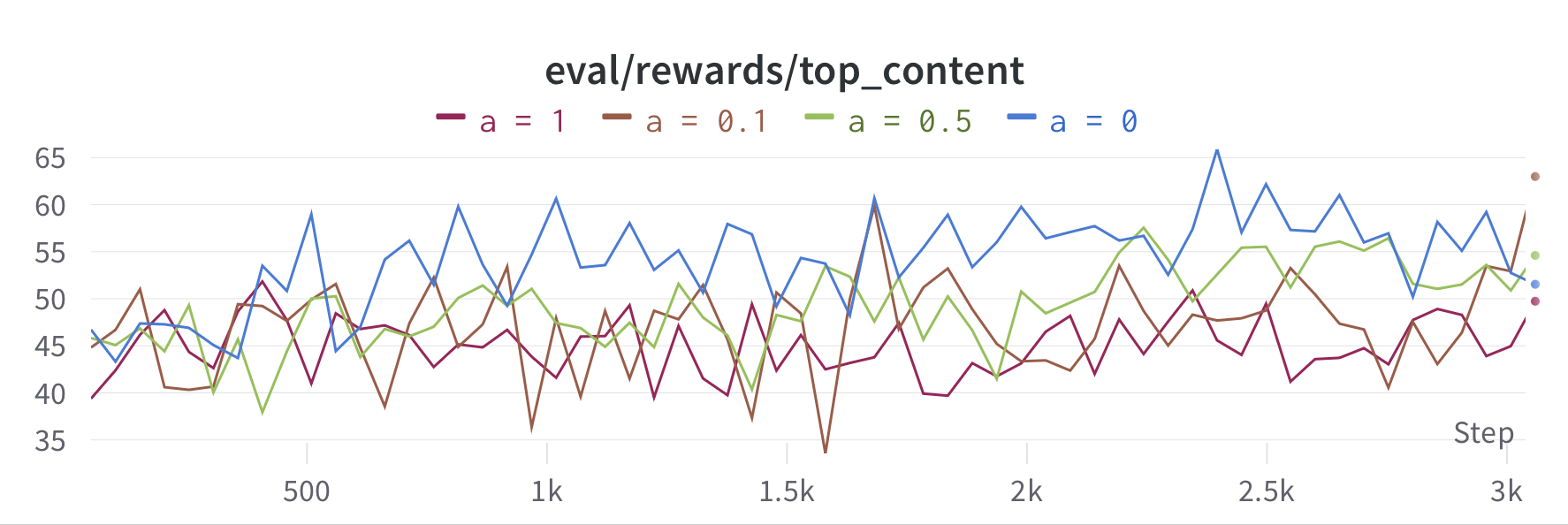}
    \includegraphics[scale=0.4]{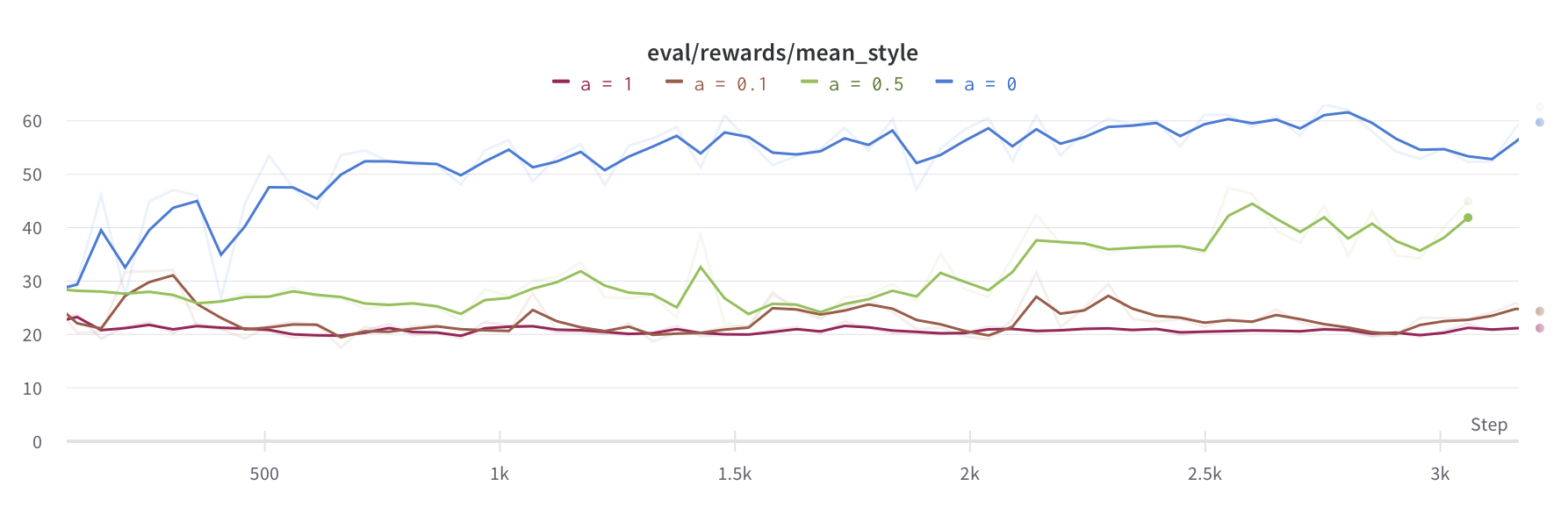}
    \includegraphics[scale=0.4]{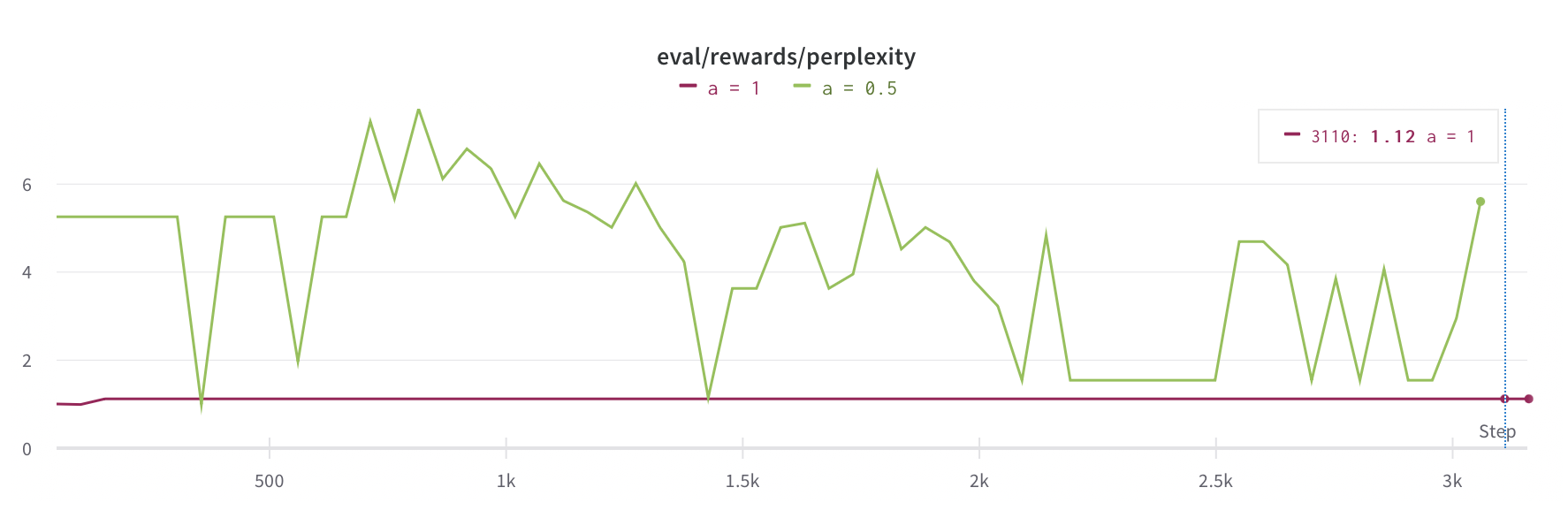}
    \caption{RLPrompt with perplexity regularization results on Yelp text style transfer with varying levels of $\alpha$. Content and style correspond to prompt performance and perplexity is our proxy for prompt interpretability.}
    \label{fig:my_label}
\end{figure}


\end{document}